\documentclass{article}

\PassOptionsToPackage{numbers, compress}{natbib}

\usepackage[final]{neurips_2024}

\usepackage{graphicx}
\usepackage{graphicx}
\usepackage{amsmath}
\usepackage{amssymb}
\usepackage{booktabs}
\usepackage{chngpage}
\usepackage{multirow}
\usepackage{cuted}
\usepackage{enumitem}
\usepackage{pifont}
\usepackage[dvipsnames]{xcolor}
\usepackage{colortbl}
\usepackage{algorithm,algorithmic}
\usepackage{amsmath}
\usepackage{comment}

\usepackage[pagebackref,breaklinks,colorlinks]{hyperref}

\usepackage[capitalize]{cleveref}
\crefname{section}{Sec.}{Secs.}
\Crefname{section}{Section}{Sections}
\Crefname{table}{Table}{Tables}
\crefname{table}{Tab.}{Tabs.}

\usepackage[utf8]{inputenc} 
\usepackage[T1]{fontenc}    
\usepackage{hyperref}       
\usepackage{url}            
\usepackage{booktabs}       
\usepackage{amsfonts}       
\usepackage{nicefrac}       
\usepackage{microtype}      
\usepackage{xcolor}         

\title{Anatomical 3D Style Transfer Enabling Efficient Federated Learning with Extremely Low Communication Costs}

\author{%
Yuto Shibata$^{1}$\thanks{This work includes contributions made during a summer internship at Preferred Networks.} \quad Yasunori Kudo$^{1}$ \quad Yohei Sugawara$^{1}$\\
$^1$Preferred Networks\\
\texttt{yuto19990715@gmail.com}\\
\texttt{\{ykudo,suga\}@preferred.jp}\\
}

\begin{document}

\maketitle

\begin{abstract}
In this study, we propose a novel federated learning (FL) approach that utilizes 3D style transfer for the multi-organ segmentation task. The multi-organ dataset, obtained by integrating multiple datasets, has high scalability and can improve generalization performance as the data volume increases. However, the heterogeneity of data owing to different clients with diverse imaging conditions and target organs can lead to severe overfitting of local models. To align models that overfit to different local datasets, existing methods require frequent communication with the central server, resulting in higher communication costs and risk of privacy leakage. 
To achieve an efficient and safe FL, we propose an Anatomical 3D Frequency Domain Generalization (A3DFDG) method for FL. A3DFDG utilizes structural information of human organs and clusters the 3D styles based on the location of organs. By mixing styles based on these clusters, it preserves the anatomical information and leads models to learn intra-organ diversity, while aligning the optimization of each local model.
Experiments indicate that our method can maintain its accuracy even in cases where the communication cost is highly limited ($=1.25\%$ of the original cost) while achieving a significant difference compared to baselines, with a higher global dice similarity coefficient score of 4.3\%.
Despite its simplicity and minimal computational overhead, these results demonstrate that our method has high practicality in real-world scenarios where low communication costs and a simple pipeline are required.
The code used in this project will be publicly available.
\end{abstract}

\section{Introduction}
\label{sec:intro}

Recently, the effectiveness of deep learning (DL) in the medical field has been demonstrated through classification and segmentation tasks~\cite{rajpurkar2017chexnet,isensee2018nnu}. 
A large amount of labeled data is required for accurate medical segmentation via DL. This requirement is critical in the field of healthcare because annotating medical images requires high levels of expertise, and collecting pixel-level labels for segmentation tasks is time-consuming and expensive. Furthermore, medical data are highly personal and confidential, making it challenging to share raw data across institutions and countries.

\begin{figure}[t]
\centering
\centerline{\includegraphics[width=1.0\linewidth]{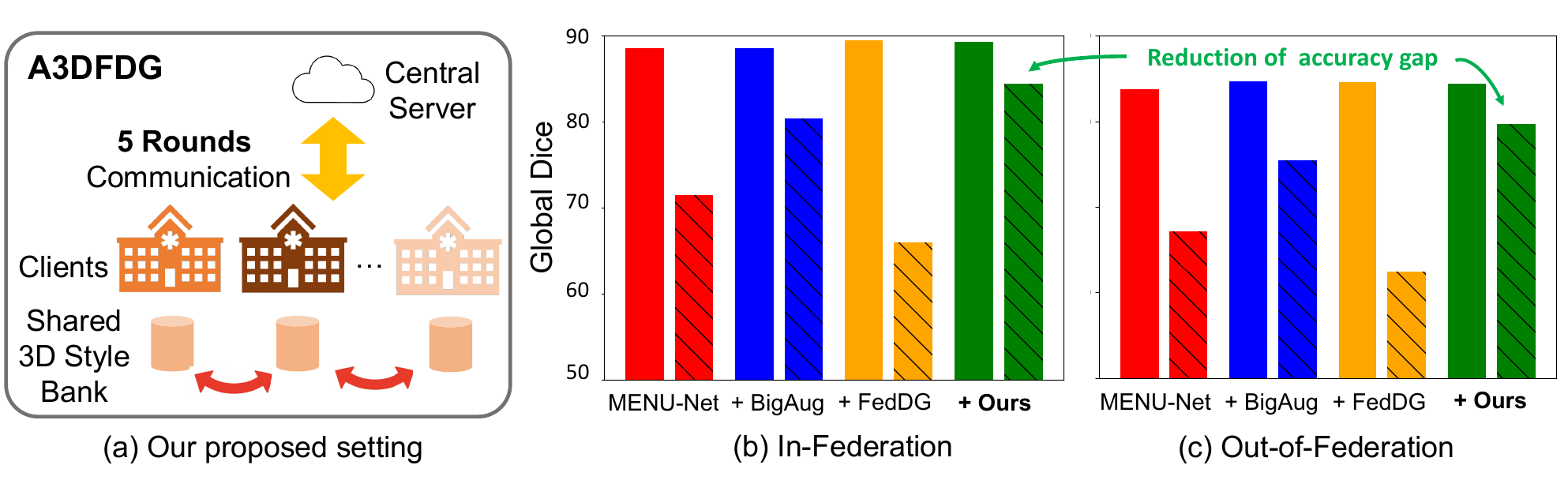}}
\vspace{-3mm}
\caption{(a) Our proposed setting with limitted communication rounds and shared style information. (b)(c) Model accuracy evaluation. The hatched bars indicate the accuracy when the number of communications is reduced from 400 to 5.}
\label{fig:acc_per_rounds}
\vspace{-5mm}
\end{figure}

In response to the aforementioned challenges, many studies have aimed to simultaneously protect patient privacy and increase the amount of data available for training by utilizing federated learning (FL)~\cite{mcmahan2017communication,wu2021federated,xu2022closing}. Under the FL scheme, distributed clients perform training using local data and upload their weights to a central server. The central server then aggregates these weights to acquire a more generalized model, whereas all local data are stored under the distributed clients~\cite{mcmahan2017communication}. 

Another approach for increasing the amount of annotated data involves combining multiple datasets targeting different organs, resulting in a multi-organ dataset~\cite{xu2023federated,Liu_2023_ICCV}. By integrating datasets, the amount of training data increases, resulting in higher accuracy than when training with individual datasets~\cite{xu2023federated}. 
However, implementing FL using these integrated datasets presents a highly challenging issue: \textbf{since each local model is optimized in different directions, more iterations are required for the global model to converge, leading to increased overall FL training time and communication cost.}
In addition to the well-known domain shift caused by image appearance variation owing to different imaging equipment and protocols~\cite{bigaug}, differences occur in the targeted organs and the imaging ranges because each dataset was prepared for different purposes. These two sources of domain shift force local models to overfit to their local datasets, significantly slowing down the overall FL training convergence. 

Fig~\ref{fig:acc_per_rounds} illustrates the inefficiency of the existing FL models by demonstrating a significant drop in accuracy with low communication costs (i.e., the number of uploads of the local models to the central server from each local client) while maintaining the total iterations. 
An in-federation setting means that test data were provided by clients included in the training data, and an out-of-federation setting means that we evaluate the accuracy against unseen clients. The accuracy under the conventional setting with 400 rounds of communications is represented by the plain bars, while the accuracy with five rounds of communications is depicted by the hatched bars.
Owing to the aforementioned variation in the optimization directions of local models caused by domain shift, existing models fail to converge and experience a significant drop in accuracy when there is a strict limitation on the number of communications with the central server.
This leads to higher operational and communication costs, increases the risk of data leakage when sharing the model, and hinders the system’s practical application.

\begin{figure}[t]
\centering
\centerline{\includegraphics[width=1.0\linewidth]{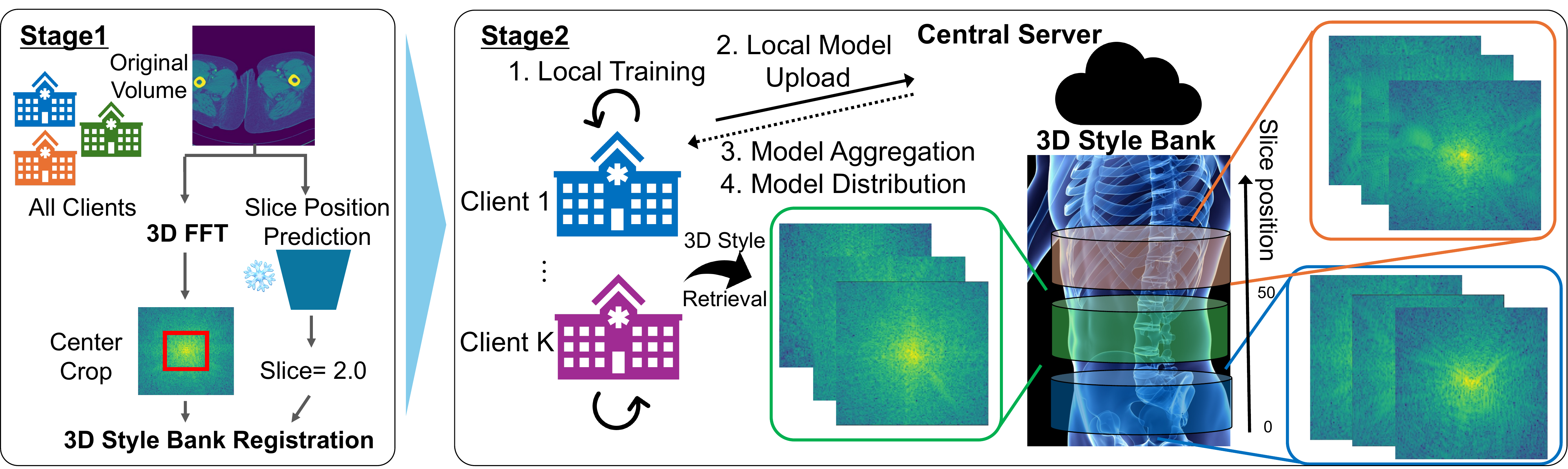}}

\caption{Overview of our A3DFDG. (Stage1) First, we calculate the 3D visual style of each client in the frequency space and store them in clusters based on the predicted slice scores (slice position). (Stage2) During training, we retrieve 3D styles from the same cluster as the samples in the minibatch and perform style transfer without losing organ information.
}
\label{fig:overview}
\end{figure}

To achieve FL with sufficient scalability and practicality, we propose a novel problem setting, \textbf{federated domain generalization with few-round communications}. Toward this goal, we propose \textbf{A3DFDG}, \textbf{A}natomical \textbf{3D} \textbf{F}requency \textbf{D}omain \textbf{G}eneralization for FL (Fig.~\ref{fig:acc_per_rounds} (a), Fig.~\ref{fig:overview}). This novel method utilizes domain generalization in the frequency space to eliminate differences between domains (clients) and resolve optimization interference among local models. 
Specifically, we develop a module that successfully extends the existing style adaptations defined in the 2D frequency domain (FDA)~\cite{yang2020fda} to 3D without sharing raw images across clients while utilizing anatomical structural information. 

Existing data augmentation methods mix multiple samples randomly, whether the mixing is done in the spatial domain~\cite{zhang2017mixup,yun2019cutmix} or in the frequency domain~\cite{yang2020fda,liu2021feddg}.
However, when dealing with multi-organ datasets, mixing styles obtained from two different organs can lead to the loss of class information contained in each style, potentially distorting the decision boundaries of the model. 
Therefore, we cluster the styles in the frequency domain using an off-the-shelf organ localization model. 
Based on these clustered 3D styles, we perform data augmentation in 3D frequency domain, while preserving anatomical information and aligning the optimization of each client's model.

To demonstrate the effectiveness of the proposed method, we conducted evaluations using six datasets under two federated learning (FL) settings: an in-federation setting and an out-of-federation setting. Our results show that even when the number of communications with the central server was significantly reduced to 1.25\% of the original setting, we maintained high accuracy comparable to that of frequent communication settings (Fig~\ref{fig:acc_per_rounds}). Conversely, existing baseline methods are unable to learn multiple organs in a balanced manner and fail to achieve convergence of the global model with limited communication. These findings demonstrate that our model is highly practical for conducting large-scale learning with fewer communications and a simpler overall pipeline.

Table 1 summarizes the communication cost and accuracy when using the recently proposed MENU-Net baseline (size: 6.3GB). Here, our method distributes the 3D style (0.23MB) among clients only once before FL training and these styles are stored at local clients without the need for redistribution during training. Also, as mentioned later, domain generalization is performed using only a part of the frequency spectrums, resulting in much smaller additional communication overhead.
\begin{table}[tbh]
    \centering
    \caption{Comparison of the trade-off between communication cost and prediction accuracy}
    \begin{tabular}{lcccc}
        \toprule
        Method & Shared Data $\times$ \# Rounds & Data traffic & DSC(\%) \\
        \midrule
        Existing Work~\cite{xu2023federated} & Model$\times$400& 2.5T & 88.49\\
        + Communication Reduction & Model$\times$5& \textbf{31.5G} & 71.42\\
        + A3DFDG & Model$\times$5+3D Style$\times$1& \textbf{31.5G} & 84.38\\
        \bottomrule
    \end{tabular}
    \label{tab:acc_and_commu}
\end{table}

Our contributions are summarized as follows.
(1) We propose the task of medical FL with low communication cost and diverse datasets;
(2) we introduce A3DFDG, a domain generalization method utilizing organ structural and low frequency information; (3) we conducted extensive experiments under various FL and communication settings, demonstrating that our simple, yet effective method outperforms existing baselines in settings with limited communication cost.

\section{Background and Related Work}
{\bf{Federated Learning.}}
In the FL framework~\cite{mcmahan2017communication}, which aims to protect the privacy of patients, distributed learning is conducted without sharing the local data among clients. Clients perform a fixed number of learning iterations using their local datasets, after which the weights of their models are aggregated on a central server. This process involves calculating the weighted average of the local models to obtain a global model~\cite{mcmahan2017communication}. The global model is then redistributed to each client for further local training, and this process is repeated until the global model converges. 

\noindent
{\bf{Domain generalization.}}
In the field of medical imaging, domain shift is a common issue owing to differences in imaging parameters and subject cohorts among hospitals. To address this problem, numerous domain adaptation and generalization methods (both supervised and unsupervised) have been proposed. However, the raw data cannot be shared in the FL framework, strictly prohibiting conventional domain adaptation/generalization approaches. For example, we cannot adopt an instance weighting strategy~\cite{wachinger2016domain,8094884} that requires the similarity scores between the source and target domains because it uses the latent features of each domain. 
Recently, some studies have addressed domain generalization using the FL scheme. For instance, ~\cite{huang2023rethinking} calculated prototypes that represent each domain in the feature space, and ~\cite{wang2023dafkd} created synthesized datasets using a generative model to train domain classifiers to obtain domain-invariant features when training local models. 
However, recent studies have revealed that the latent features of trained models risk privacy data leakage~\cite{singh2021disco} even under collaborative training schemes~\cite{he2019model,yang2022measuring}, making these approaches unsuitable in the medical field where strong privacy protections is required.  
Other studies calculated styles based on FDA~\cite{yang2020fda} to obtain the domain knowledge of each client because the amplitude information of lower frequency bands cannot be used for original image reconstruction without higher frequency and phase components~\cite{shenaj2023learning,liu2021feddg}. However, to the best of our knowledge, no studies have ever implemented these frequency-based approaches toward 3D medical segmentation tasks with diverse multi-organ datasets. 

\noindent
{\bf{Multi-organ datasets.}}
Many existing studies that perform learning by combining multiple medical datasets focus on addressing the issue of partial labels in the integrated dataset~\cite{shi2021marginal, xu2023federated}. \cite{shi2021marginal} proposed the exclusion and marginal loss to calculate additional supervision with partial label and \cite{xu2023federated} tackled the partial label problem under the FL scheme by separating the encoder into sub-encoders to prevent expert models, which undergo supervision with labels, from losing their knowledge by averaging their weights with other non-expert models. Note that this paper addresses the domain shift that arises from merging multiple datasets; thus, handling partial labels falls outside the scope of our proposed method.

\noindent
{\bf{Low communication FL.}}
FL requires many communication rounds between a central server and its clients to achieve high accuracy, increasing computational costs and the risk of privacy leakage~\cite{zhang2022dense,feng2023learning,park2021few,song2022resfed}. Furthermore, frequent communication complicates the FL pipeline, making it challenging to handle confidential local data and tmodels. Existing studies have reduced communication costs by decreasing the size of trainable models or sharing the residuals of updates~\cite{feng2023learning,song2022resfed}. Regarding the reduction in communication frequency, a recent study addressed few-round FL~\cite{zhang2022dense,park2021few}. However, DENSE~\cite{zhang2022dense} assumes the same task among clients, making it difficult to apply to multi-organ FL. 
Additionally, these methods have been implemented for relatively simple recognition tasks such as CIFAR-10/100, and they have not been used for 3D medical image segmentation, which requires both high-level and fine-grained visual understanding.

\begin{figure*}[t]
\centering
\centerline{\includegraphics[width=1.0\linewidth]{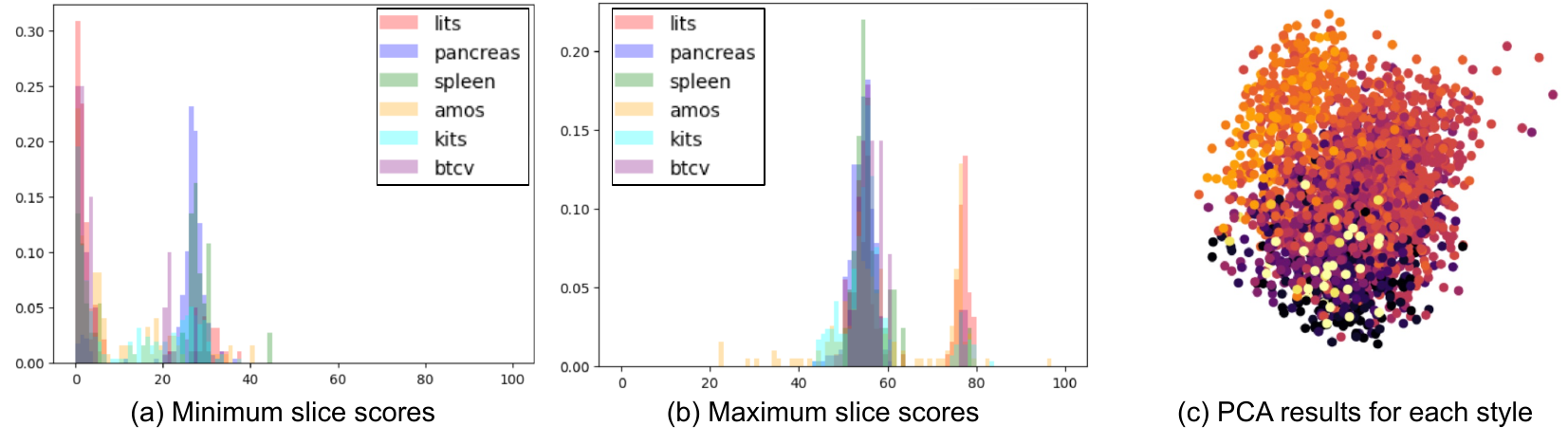}}
\vspace{-3mm}
\caption{(a)(b) Distribution of the predicted slice scores (= predicted slice position at which images are captured) across six datasets. (c) The distribution of extracted style. Here, color indicates its slice score and PCA is implemented for visualization.}
\label{fig:height_info}
\vspace{-2mm}
\end{figure*}

\section{Method}
Fig.~\ref{fig:overview} shows the overview of our method. Our proposed framework consists of three parts designed to share the domain information of local datasets while avoiding raw feature leakage and losing anatomical information. 
Secs.~\ref{sec:style_registration} and \ref{sec:anatomical_registration} describe the style calculation and the style clustering based on organ positions. Sec.~\ref{sec:style_share_training} then explains the training of our models based on these registered style banks. 
Algorithm~\ref{alg:bank} presents the detailed steps of our proposed method. 

\noindent
\textbf{Preliminaries.} In this study, K is the number of clients with local datasets. They are expressed as $D = \{D_1, D_2, D_3, \ldots, D_K\}$. Each client contains its local confidential data $D_k = \left\{ (x_i^k, y_i^k) \right\}_{i=1}^{N^k}$ while $N_k$, $x^k_i$, and $y^k_i$ indicate the size of local data, $i$-th CT volume, and labels in the $k$-th dataset, respectively.

\subsection{3D style calculation}
\label{sec:style_registration}
Previous studies~\cite{liu2021feddg,shenaj2023learning,yang2020fda} have proved that transferring distribution information in 2D frequency space across clients is effective for domain generalization in the FL training. Toward 3D medical segmentation, our proposed method uses the outputs of a 3D Fourier transform applied to a volume as a style representing each client's domain.
First, we respace (up/down-sample) each voxel because the difference in imaging space results in frequency resolution discrepancy when implementing Fourier transformation. We then extract volumes of the same size as used during training and prediction from each voxel at a consistent height interval, thereby creating a 3D style bank keyed by height (the motivation for registration by height is described in Sec.~\ref{sec:anatomical_registration}). The amplitude and phase components of the Fourier transform are denoted as $\mathcal{F^A}$ and $\mathcal{F^P}$, respectively. The style of each local volume $x_i^k$ is expressed as follows:


\begin{align}
s^k_i(u,v,t) &= \mathcal{F^A}(x_i^k)(u, v, t) = |\sum_{h=0}^{H-1} \sum_{w=0}^{W-1} \sum_{d=0}^{D-1} x_i^k(h, w, d) e^{-j 2\pi \left( \frac{hu}{H} + \frac{wv}{W} + \frac{dt}{D} \right)}|
\label{3d_fourier}
\end{align}
where $H, W,$ and $D$ indicate the height, width, and depth of cropped volume, respectively.
To preserve the privacy of the patients, we center-crop each Fourier 3D representation, and only the amplitude information in the low-frequency band is registered in the style bank, thus making it impossible to reconstruct the original volume. Additionally, these styles are extracted only from the training data to prevent potential leakage from test data. Following the previous work~\cite{liu2021feddg,yang2020fda}, we calculate weighted sums between different frequency styles to represent continuous and diverse 3D domain information (see Sec.~\ref{sec:style_share_training} for detail).

\begin{figure*}[t]
\centering
\centerline{\includegraphics[width=1.0\linewidth]{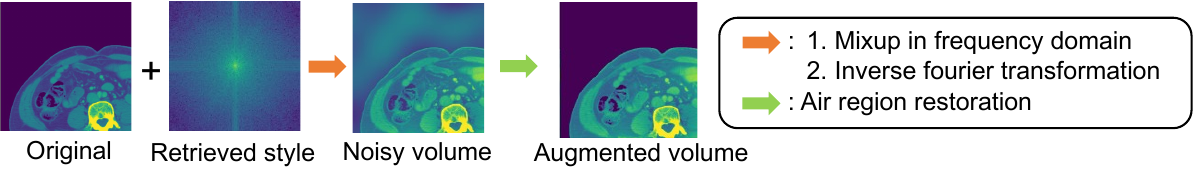}}
\vspace{-4mm}
\caption{Data augmentation in the frequency domain and subsequent post-processing}
\label{fig:data_aug_visualizaton}
\vspace{-2mm}
\end{figure*}

\subsection{Anatomical position registration}
\label{sec:anatomical_registration}
Our dataset is a multi-organ dataset created by merging data from multiple clients, each featuring different organs. As a result, while the dataset is large in scale, it covers a wide range of anatomical locations. To investigate the statistics regarding this aspect, we used a pretrained off-the-shelf body part regressor~\cite{yan2018unsupervised}. This model outputs slice scores, relative height of the imaging location, where the pelvis and head are set to 0 and 100, respectively. Fig.~\ref{fig:height_info}(a) and (b) show the distribution of the maximum and minimum slice scores of the ranges spanned by each volume in the datasets. We can see that different datasets feature different organs, resulting in variations in the imaging locations and potential domain shifts among clients.

To examine the relationship between organ position and domain information defined in the frequency space, we performed dimensionality reduction on the 3D Frequency styles described in Sec.~\ref{sec:style_registration} using PCA and visualized the results (Fig.~\ref{fig:height_info} (c)). Here, different colors correspond to different estimated slice scores. This visualization demonstrates that styles with the same color tend to cluster together, indicating that our 3D styles encapsulate information related to the position of the organs.

However, with these frequency styles that contain organ position information, randomly mixing the two styles for data augmentation as has been done previously in the spatial or frequency domains \cite{zhang2017mixup,yun2019cutmix,liu2021feddg} can result in the loss of crucial organ information while distorting the decision boundaries of the model. In other words, by mixing these styles during training, the model is optimized to make predictions without relying on the slice position information of the organs, even though this information is actually beneficial for organ identification. Therefore, in this study, we first predict organ locations using a pretrained organ locator model~\cite{yan2018unsupervised} and cluster each style based on predetermined binning thresholds (see 3D Style Bank in Fig.~\ref{fig:overview}). For each cluster, we then perform data augmentation in the frequency domain. This approach forces models to learn intra-organ diversity without imparting biases related to incorrect low-frequency information to the model.
Nnote that each client calculates the slice scores for their data only once before the FL training, and the body position estimator~\cite{yan2018unsupervised} is lightweight, resulting in a minimal additional computation cost for the client.

\begin{algorithm}
\caption{3D Style Bank Registration}
\begin{algorithmic}[1]
\renewcommand{\algorithmicrequire}{\textbf{Input:}}
\renewcommand{\algorithmicensure}{\textbf{Output:}}
\REQUIRE Off-the-shelf body part regressor $\phi(w)$, clients $k \in \mathcal{K}$ with local dataset $D_k$, pre-defined slice score bin size $z_{bin}$ and volume size width $w$, height $h$, and depth $d$
\ENSURE  Unified 3D Style Bank $B$
  \FOR {client $k \in \mathcal{K}$}
  \FOR {volume $v \in \mathcal{D}_k$}
  \STATE Predict the maximum and minimum slice scores with $\phi(w)$ and calculate volume height\\ $z_{length}=z_{max}-z_{min}$.
  \STATE Random crop sub-volume $v$ of size $w, h$ and $d$.
  \STATE Calculate the corresponding slice score $z'$ based on $z_{min}$, $z_{length}$, and stride size (z-axis) used for cropping.
  \STATE Calculate 3D style $s$ based on Eq.~\ref{3d_fourier} and register it for the style bank $B$.
  \begin{equation}
  B[k][z'//z_{bin}].\rm{append}(\rm{CenterCrop}(s))
  \end{equation}
  \ENDFOR
  \ENDFOR \\
 \STATE Distribute $B$ across all clients and start FL training utilizing fourier domain generalization (Eq.~\ref{eqs:style_mixup}).
\end{algorithmic}
\vspace{-1.5mm}
\label{alg:bank}
\end{algorithm}


\subsection{FL training with 3D style bank}
\label{sec:style_share_training}
Suppose we are training the $k$-th local model $f_k^t(x; \theta^t_k)$ in round $t$ with the local $i$-th data $x^k_i$, where $\theta^t_k$ denotes the $k$-th local model. In each iteration, the local model randomly selects another client $k'$ and retrieves a target 3D style $s_{target}$ that has a similar slice score with the cropped local volume from the precomputed style bank $B[k'][z_{x^k_i}//z_{bin}]$ randomly. 
During training, the slice score of each cropped sample for style retrieval is calculated based on the maximum/minimum slice score of the original volume and the size of the stride in the z-direction similar to the style bank registration (see Algorithm~\ref{alg:bank}).
Subsequently, the two styles are mixed in the frequency space as follows:
\begin{equation}
s' = \alpha\mathcal{F^A}(x^k_i)+ (1-\alpha)s_{target}
\end{equation}
\begin{equation}
\label{eqs:style_mixup}
{x^k_i}' = \mathcal{F}^{-1}\left( [\left(M_{\beta} \circ s'\right) + \left((1 - M_{\beta}) \circ \mathcal{F^A}(x^k_i)\right), \mathcal{F}^P(x^k_i) \right])
\end{equation}
where $\mathcal{F}^{-1}$ is the inverse Fourier Transform, $\alpha$ is the hyperparameter for this MixUp operation, and $M$ is a mask whose value is one at the predefined center region; otherwise, it is zero.
\begin{equation}
M_{\beta}(h, w, d) = \mathbf{1}_{(h,w, d)\in[\![-\beta_h H:\beta_h H], [-\beta_w W:\beta_w W], [-\beta_d D:\beta_d D]\!]}
\end{equation}
where $\beta$ is a hyperparameter that controls the extent of the style transfer. Note that since $\beta$ is very small (Sec.~\ref{sec:implementation_detail}), the communication cost for sharing these styles is limited (Table~\ref{tab:acc_and_commu}).

Inverse Fourier Transformation after style mixing results in artifacts in the external regions of the body as shown in Fig~\ref{fig:data_aug_visualizaton}. 
Therefore, we preserve anatomical information, such as the distance to the contour of the body using a threshold-based air mask for post-refinement. We use air threshold $\tau_{air}=-200$ to filter air pixels and those pixels are filled with the original value after style transformation.

Each time local training is completed, we obtain the global model by calculating the weighted average of each local model based on their dataset sizes following common FL implementation~\cite{mcmahan2017communication,xu2023federated}. 
\begin{equation}
\theta^{t+1} = \sum_{k} \frac{|D_k|}{\sum_{j} |D_j|} \cdot \theta^t_k.
\end{equation}

\section{Experimental Settings}
\noindent
{\bf{Datasets and Preprocessing.}}
Following previous work~\cite{xu2023federated}, we used six (multi-) organ segmentation datasets: 1) the liver tumor segmentation challenge~\cite{bilic2023liver} (LiTs), 2) kidney tumor segmentation
challenge (KiTS)~\cite{heller2019kits19,heller2021statekits}, 3) pancreas, 4)
spleen segmentation datasets in medical segmentation decathlon challenge\cite{simpson2019large,antonelli2022medical}, 5)
multi-modal abdominal multi-organ segmentation challenge
(AMOS)\cite{ji2022amos}, and 6) multi-atlas labeling beyond the
cranial vault challenge (BTCV)~\cite{landman2015miccai} datasets. These datasets contain 131, 210, 281, 41, 200, and 30 volumes, respectively. For more detailed information about these datasets please refer to the prior work~\cite{xu2023federated}. 
We also implemented the same preprocessing as in~\cite{xu2023federated} for downsampling and pixel normalization with clipping. 

\noindent
{\bf{Baselines.}}
We compared our proposed model with the following baseline models: (i) FedAvg~\cite{mcmahan2017communication}, the original work on FL that calculates the average of weights after local training; (ii) MENU-Net~\cite{xu2023federated}, which separates the encoder into multiple sub-encoders to prevent model optimization interference during the global model update; (iii) BigAug~\cite{bigaug}, which utilizes a set of heavy augmentations to generalize the model towards unknown domains; (iv) FedDG~\cite{liu2021feddg}, the closest research to our work, and it achieves highly generalized Federated Learning while preserving privacy by center-cropped frequency spectrums. The 3D organ datasets we handle are difficult to collect, and some clients have limited local data (e.g., the spleen segmentation dataset~\cite{antonelli2022medical} contains only 24 training samples). Therefore, instead of adopting the meta-learning approach proposed in the FedDG paper, we adopted only the data augmentation part toward the multi-source domain based on continuous frequency space interpolation. 
Regarding (iii) (iv), MENU-Net was adopted as the network architecture. In addition, (i) uses the same 3D convolutional layers as MENU-Net except for sub-encoders. 
For more detailed hyperparameter settings, please refer to the Supplementary Material. 

\noindent
{\bf{Evaluation Metrics.}}
We calculated the average dice similarity coefficient score (DSC) and average symmetric surface distance (ASD) for each organ. The macro average across clients was calculated in an in-federation setting. Also, the global accuracy was calculated using the macro average across all organs.

\noindent
{\bf{Implementation Details.}}
\label{sec:implementation_detail}
We adopted the MENU-Net architecture for the model of our proposed method and trained our model with dice, cross-entropy, and marginal and exclusion loss functions following~\cite{xu2023federated}.
Regarding the hyperparameters, $\alpha$ is randomly sampled within $[0.0, 1.0]$, and we set $\beta_w, \beta_h, $, and $\beta_d$ to 0.01, 0.01, and 0.05, respectively. These hyperparameters are used for both of our method and FedDG~\cite{liu2021feddg}. We used 10\% and 30\% of each local data for validation and testing while all images in BTCV were used for out-of-federation testing. 
The training and testing batch sizes were set to four and two, respectively.
The learning rate was set to 0.01 for 400 communication rounds setting and 0.001 for 5 rounds settings to stabilize the training processes. 

\begin{table*}[t]
\centering
\caption{Dice similarity coefficient (DSC) scores in the in-federation setting.}
{ 
\begin{tabular}{lccccccc}
\toprule
 &  & \multicolumn{5}{c}{DSC (\%)} \\ 
 \cmidrule{3-8}
 \multirow{1}{*}{Model} & \multirow{1}{*}{\# Rounds} & Liver & Kidney & Pancreas & Spleen & Gallbladder & Global \\ 
 \midrule
FedAvg~\cite{mcmahan2017communication} & 400 & \textbf{94.95} & 94.62 & 81.40 & 92.13 & 80.24 & 88.67 \\ 
 & 5 & 91.29 & 90.28 & 58.75 & 75.07 & 43.46 & 71.77 \\ 
\midrule
MENU-Net~\cite{xu2023federated} & 400 & 94.43 & 94.85 & 81.97 & 92.60 & 78.6 & 88.49 \\ 
 & 5 & 92.29 & 91.64 & 76.44 & 28.92 & 67.81 & 71.42\\ 
\midrule
MENU-Net & 400 & 93.94 & 94.25 & \textbf{82.08} & 91.70 & 80.62 & 88.52 \\ 
+ BigAug~\cite{xu2023federated,zhang2020generalizing} & 5 & 91.19 & 90.00 & 75.12 & 80.63 & 64.43 & 80.28 \\ 
\midrule
MENU-Net & 400 & 94.39 & \textbf{94.87} & 81.77 & \textbf{93.18} & \textbf{82.77} & \textbf{89.40} \\ 
 + FedDG~\cite{xu2023federated,liu2021feddg} & 5 & \textbf{92.52} & 80.81 & \textbf{77.79} & 8.51 & 69.89 & 65.90 \\ 
 \midrule
\rowcolor[rgb]{0.9, 0.9, 0.9}
Ours & 400 & 94.57 & 94.61 & 81.64 & 93.02 & 82.21 & 89.21 \\ 
\rowcolor[rgb]{0.9, 0.9, 0.9}
 \rowcolor[rgb]{0.9, 0.9, 0.9}
 & 5 & 92.51 & \textbf{92.59} & 77.17 & \textbf{89.23} & \textbf{70.42} & \textbf{84.38}\\ 
\hline
\end{tabular}
} 
\label{tab:quanti_in_fed}
\vspace{-1mm}
\end{table*}

\begin{table*}[t]
\centering
\caption{DSC scores in the out-of-federation setting.}
{ 
\begin{tabular}{lccccccc}
\toprule
 &  & \multicolumn{5}{c}{DSC (\%)} \\ 
 \cmidrule{3-8}
 \multirow{1}{*}{Model} & \multirow{1}{*}{\# Rounds} & Liver & Kidney & Pancreas & Spleen & Gallbladder & Global \\ 
 \midrule
FedAvg~\cite{mcmahan2017communication} & 400 & \textbf{95.03} & \textbf{88.57} & 79.10 & 90.99 & 67.84 & 84.31\\ 
 & 5 & 92.20 & 87.89 & 53.07 & 66.59 & 32.42 & 66.43 \\ 
\midrule
MENU-Net~\cite{xu2023federated} & 400 & 94.83 & 87.87 & 77.85 & 89.19 & \textbf{68.73} & 83.69 \\ 
 & 5 & 94.41 & 87.65 & 74.45 & 33.23 & 45.78 & 67.10 \\ 
\midrule
MENU-Net & 400 & 94.57 & 87.77 & 79.58 & 90.72 & 70.70 & \textbf{84.67} \\ 
+BigAug~\cite{xu2023federated,zhang2020generalizing} & 5 & 93.98 & 85.44 & 72.20 & 81.04 & 44.13 & 75.36\\ 
\midrule
MENU-Net & 400 & 94.95 & 88.12 & 78.77 & \textbf{92.27} & 68.59 & 84.54 \\ 
+ FedDG~\cite{xu2023federated,liu2021feddg} & 5 & 94.15 & 78.40 & 75.92 & 6.84 & \textbf{56.59} & 62.38 \\ 
\midrule
\rowcolor[rgb]{0.9, 0.9, 0.9}
Ours & 400 &94.76&87.68&\textbf{79.65}&91.67&68.21& 84.39\\ 
\rowcolor[rgb]{0.9, 0.9, 0.9}
 \rowcolor[rgb]{0.9, 0.9, 0.9}
 & 5 & \textbf{94.64} & \textbf{89.35} & \textbf{75.98} & \textbf{88.79} &49.56 &\textbf{79.66}\\ 
\midrule
\end{tabular}
} 
\label{tab:quanti_out_fed}
\vspace{-1mm}
\end{table*}

\begin{table*}[th]
\centering
\caption{ASD scores in the out-of-federation setting.}
\begin{tabular}{lccccccc}
\toprule
 &  & \multicolumn{5}{c}{ASD (mm)} \\ 
 \cmidrule{3-8} 
 \multirow{1}{*}{Model}& \multirow{1}{*}{\# Rounds} & Liver & Kidney & Pancreas & Spleen & Gallbladder & Global \\ \midrule
FedAvg~\cite{mcmahan2017communication} & 400 & 2.57 & \textbf{4.27} & \textbf{1.98} & 1.67 & 2.60 & 2.62\\ 
 & 5 & 3.91 & 6.08 & 38.15 & 5.18 & 11.18 & 12.90\\ 
\midrule
MENU-Net~\cite{xu2023federated} & 400 & \textbf{2.53} & 4.74 & 2.66 & 2.43 & 2.21 & 2.91 \\ 
 & 5 & 3.33 & 6.41 & 4.70 & 166.84 & \textbf{2.55} & 36.77\\ 
\midrule
MENU-Net 
 & 400 & 2.83 & 4.95 & 2.23 & 2.22 & 2.00 & 2.84 \\ 
+ BigAug~\cite{xu2023federated,zhang2020generalizing} & 5 & 2.74 & 7.64 & 12.25 & \textbf{3.61} & 5.85 & 6.42\\ 
\midrule

MENU-Net 
 & 400 & 2.60 & 4.90 & 2.69 & \textbf{1.40} & \textbf{1.22} & \textbf{2.56}\\ 
+ FedDG~\cite{xu2023federated,liu2021feddg}& 5 & 2.66 & 9.24 & 3.22 & 26.16 & 8.41& 9.94\\ 
\midrule

\rowcolor[rgb]{0.9, 0.9, 0.9}
Ours & 400 & 2.71 & 5.31 & 2.49 & 2.05 & 3.99 & 3.31\\ 
\rowcolor[rgb]{0.9, 0.9, 0.9}
 & 5 & \textbf{2.49} & \textbf{4.69} & \textbf{2.59} & 4.34 & 3.91 & \textbf{3.60}\\ 
\midrule
\end{tabular}
\label{tab:asd_out_fed}
\vspace{-2mm}
\end{table*}

\section{Experimental Results.}
\label{sec:experimental_results}
To evaluate the efficacy of our model and the other baselines, we trained and evaluated them with two communication frequencies under two distinct settings: (i) an in-federation setting and (ii) an out-of-federation setting.

\subsection{Comparison with Other Baselines}
\label{quantitative_results}
Table~\ref{tab:quanti_in_fed} presents a quantitative comparison in the in-federation setting. We can see that the methods using domain generalization in frequency space recorded high accuracy under the frequent communication setting (Rounds$=400$). However,  only the proposed method is able to maintain high accuracy even when the number of communications with the central server is significantly reduced to 1.25 \% of the original cost (Rounds$=5$). Also, it can be observed that many previous studies face challenges in maintaining accuracy for all organs when the number of communications is restricted. For example, while FedDG~\cite{liu2021feddg} achieves high accuracy for all organs with frequent communications, in the limited communication setting, the training for the spleen and kidney has not converged, resulting in significantly lower accuracy for these organs.

\begin{figure*}[t]
\centering
\centerline{\includegraphics[width=1.0\linewidth]{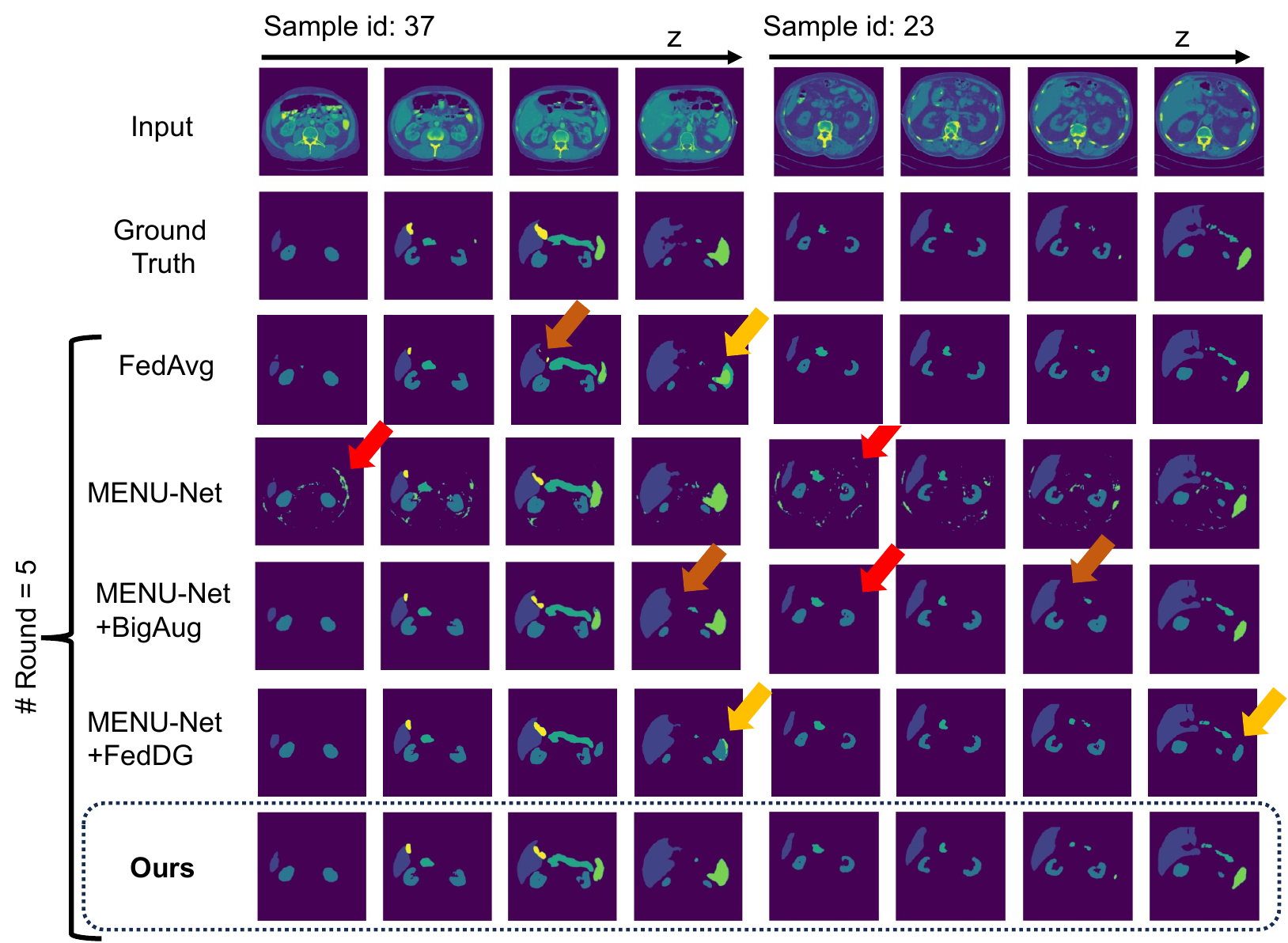}}
\vspace{-3mm}
\caption{Qualitative results in the out-of-federation setting.}
\label{fig:quali_out}
\vspace{-2mm}
\end{figure*}

Table~\ref{tab:quanti_out_fed} presents the quantitative results in the out-of-federation setting. 
Similar to the in-federation setting, accuracy reduction is limited ($< 5\%$) even when the number of rounds is significantly reduced while other baseline methods significantly degraded their accuracy (-18\%, -17\%, -8\%, -22\%), indicating that our domain generalization method enables efficient FL training while reducing the optimization interference among local models. 

Table~\ref{tab:asd_out_fed} shows the ASD scores in the out-of-federation setting. The accuracy of our proposed method is significantly higher than that of existing methods in realistic settings under low communication costs while other models have very unstable training processes and fail to converge. For the results in the in-federation setting, please refer to our supplementary material.

These results suggest the following. First, when communication rounds are limited, performing heavy data augmentation (ours, BigAug~\cite{bigaug}) achieves efficient model aggregation. Alternatively, when random mixup is applied as in FedDG~\cite{liu2021feddg}, incorrect biases can be introduced into the local models, leading to lower accuracy for some organs compared to when no data augmentation is applied.

\subsection{Qualitative results}
Fig~\ref{fig:quali_out} shows the qualitative results in an out-of-federation setting. These results included two patients, and we displayed the results by slicing at equal intervals in the z-direction. The existing methods~\cite{mcmahan2017communication,xu2023federated,bigaug,liu2021feddg} produce many false positives (red arrows), false negatives (brown arrows), and misclassification (yellow arrows) in a low communication setting. By contrast, our proposed method maintains high accuracy even when the number of communications is restricted.

\subsection{Ablation study}
We investigated the effects of our technical contributions, including position-based style clustering (slice score matching) and post-processing with an air mask. Table~\ref{tab:ablation} presents the accuracy in the in-federation and the out-of-federation setting respectively. Based on these results, we can see that both our proposed modules contributed to improving the estimation accuracy compared with the scores in Table~\ref{tab:quanti_in_fed}, specifically in a low communication setting.
\label{ablation_study}

\subsection{Model-Agnostic Efficacy of our method}
Tables~\ref{tab:model_agnostic_in_fe} demonstrates the global accuracy in both settings when we apply our method for the FedAvg~\cite{mcmahan2017communication} model. We can observe that our method significantly improved accuracy at lower communication costs, demonstrating that it has a minimal dependency on model architecture choice and a high potential to be utilized as a plug-in function for various model architectures.



\begin{table}[tb]
\centering
\caption{Ablation study}
\begin{tabular}{lccc}
\toprule
Model & {\#Rounds} & {In-Fed DSC (\%)} & {Out-of-Fed DSC (\%)}\\ \midrule
Ours & 400 & 88.69 (-0.52) & 83.51 (-0.88)\\ 
w/o slice score matching & 5 & 73.01 (-11.37) & 65.97 (-13.68)\\ 
\midrule
Ours & 400 & 87.90 (-1.3) & 84.69 (+0.30) \\ 
w/o contour preservation & 5 & 83.01 (-1.37) & 78.16  (-1.5)\\
\midrule
\end{tabular}
\label{tab:ablation}
\vspace{-2mm}
\end{table}

\begin{table}[tb]
\centering
\caption{Model-agnostic efficacy}
\begin{tabular}{lccc}
\toprule
Model & {\# Rounds} & {In-Fed DSC(\%)} & {Out-of-Fed DSC(\%)} \\ 
\midrule
FedAvg~\cite{mcmahan2017communication} & 400 & 88.67 & 84.31\\ 
 & 5 & 71.77 & 66.43\\ 
\midrule
FedAvg~\cite{mcmahan2017communication}+A3DFDG & 400 &  88.62 (-0.05) & 84.07 (-0.28)\\ 
& 5 &  73.71 (+1.94) & 68.56 (+2.13)\\
\bottomrule
\end{tabular}
\label{tab:model_agnostic_in_fe}
\vspace{-2mm}
\end{table}

\section{Discussion and Limitations}
Although our proposed method significantly improves accuracy in settings with limited communication costs, the improvement margin is limited in scenarios where frequent communication is possible. 
Moreover, one of the current main limitations of our framework is that each client needs to calculate the height of local volumes using a pre-trained organ position estimator beforehand. In future work, we plan to address this by using segmentation predictions to determine the volume occupied by each organ and dynamically calculating the corresponding slice score on the fly.

\section{Conclusion}
This paper propose A3DFDG, an Anatomical 3D Frequency Domain Generalization  method to achieve efficient FL for a heterogeneous multi-organ dataset.
Compared with existing methods that randomly sample and mix styles, the proposed method utilizes 3D styles clustered based on the organ location. This approach enables domain generalization without compromising anatomical information and forces models to learn intra-organ diversity. Despite its simplicity and minimal computational overhead, our method maintains accuracy with restricted communication frequency while existing methods significantly decrease in accuracy or fail to converge. 
We believe that this work offers a new possibility for highly practical large-scale FL with limited communication costs and diverse data.

\noindent 
\textbf{Acknowledgements.}
We would like to express our gratitude to the medical imaging team at Preferred Networks for the valuable discussions and helpful feedbacks, and to the cluster team for enabling the execution of numerous experiments.

\newpage

{\small
\bibliographystyle{ieee_fullname}
\bibliography{ref}
}

\end{document}


\title{Anatomical Three-Dimensional Style Transfer Enabling Efficient Federated Learning with Extremely Low Communication Costs 
\\ \emph{Supplementary Materials}
}

\maketitle

\section{Detailed information about hyperparameters}
\label{sec:hyper_parameters}
The training and testing batch sizes are set to four and two respectively on four NVIDIA V100 GPU. Regarding BigAug implementation, we added Gaussian noise and changed the scale, zoom, rotation, contrast, offset, and smoothness of each volume randomly.

\section{ASD scores in the in-federation setting}
In the main paper, due to space constraints, we included the ASD scores for the out-of-federation setting. Here, we present the ASD accuracy for the in-federation setting in Tab~\ref{tab:asd_in_fed}. It can be observed that the proposed method outperforms existing baselines across various organs and communication settings. Additionally, in settings where communication cost is restricted, existing methods often show unstable accuracy and fail to converge during training for smaller and more difficult-to-identify organs. In contrast, the proposed method consistently achieves high accuracy.

\begin{table*}[htbp]
\centering
\caption{ASD scores in the in-federation setting.}
\begin{tabular}{l|c|c|c|c|c|c|c}
\hline
\multirow{2}{*}{Model} & \multirow{2}{*}{\# Rounds} & \multicolumn{5}{c}{ASD} \\ \cline{3-8} &                                     & Liver & Kidney & Pancreas & Spleen & Gallbladder & Global \\ \hline
FedAvg~\cite{mcmahan2017communication} & 400 & \textbf{1.80} & \textbf{0.91} & 1.75 & 1.14 & 1.23 & 1.36\\ 
 & 5 & 5.51 & 4.07 & 34.74 & 3.28 & 5.42 & 10.61\\ 
\hline
MENU-Net~\cite{xu2023federated} & 400 & 2.03 & 1.03 & 1.67 & 1.15 & 1.11 & 1.40 \\ 
 & 5 & 4.98 & 3.28 & 4.97 & 161.98 & \textbf{2.22} & 35.49\\ 
\hline
MENU-Net 
 & 400 & 2.52 & 1.26 & \textbf{1.66} & 1.60 & 2.24 & 1.86 \\ 
+ BigAug~\cite{xu2023federated,zhang2020generalizing} & 5 &  3.02 &  4.81 & 10.83 & \textbf{2.43} & 3.21 & 4.86\\ 
\hline
MENU-Net 
 & 400 & 2.30 & 0.96 & 1.81 & 0.81 & 1.24 & 1.43\\ 
+ FedDG~\cite{liu2021feddg} & 5 & 3.62 & 6.97 & 2.60 & 17.50 & 9.52 & 8.04\\ 
\hline
\rowcolor[rgb]{0.9, 0.9, 0.9}
Ours & 400 & 1.94 & 1.03 & 1.84 & \textbf{0.76} & \textbf{1.05} & \textbf{1.32}\\ 
\rowcolor[rgb]{0.9, 0.9, 0.9}
 & 5 &  \textbf{2.84} & \textbf{1.56} &  \textbf{2.93}  &  7.69 &  2.75& \textbf{3.55}\\ 
\hline
\end{tabular}
\label{tab:asd_in_fed}
\end{table*}

{\small
\bibliographystyle{splncs04}
\bibliography{ref}
}